\documentclass[letterpaper, 10 pt, conference]{ieeeconf}  

\IEEEoverridecommandlockouts                              

\usepackage{amsmath} 
\usepackage{amssymb}  

\usepackage{graphicx} 
\usepackage{subcaption}
\usepackage{multirow}
\usepackage[table]{xcolor}
\usepackage{makecell}
\let\labelindent\relax
\usepackage[shortlabels]{enumitem}
\usepackage{hyperref}
\usepackage{algorithm}
\usepackage{algcompatible}
\usepackage{bm}

\title{\LARGE \bf
Meta-Reinforcement Learning for Robotic Industrial Insertion Tasks
}

\author{Gerrit Schoettler$^{1}$, Ashvin Nair$^{2}$, Juan Aparicio Ojea$^{1}$,\\ Sergey Levine$^{2}$, Eugen Solowjow$^{1}$
\thanks{$^{1}$Siemens
        {\tt\small \{gerrit.schoettler, juan.aparicio, eugen.solowjow\}@siemens.com}}%
\thanks{$^{2}$UC Berkeley
        {\tt\small \{nair, svlevine\}@eecs.berkeley.edu}}%
}

\begin{document}

\maketitle
\thispagestyle{empty}
\pagestyle{empty}

\begin{abstract}
Robotic insertion tasks are characterized by contact and friction mechanics, making them challenging for conventional feedback control methods due to unmodeled physical effects.
Reinforcement learning (RL) is a promising approach for learning control policies in such settings.
However, RL can be unsafe during exploration and might require a large amount of real-world training data, which is expensive to collect.
In this paper, we study how to use meta-reinforcement learning to solve the bulk of the problem in simulation by solving a family of simulated industrial insertion tasks and then adapt policies quickly in the real world.
We demonstrate our approach by training an agent to successfully perform challenging real-world insertion tasks using less than 20 trials of real-world experience.
\end{abstract}

\section{Introduction}\label{sec:introduction}
How can we embed prior knowledge into robot control systems? For simple tasks, an engineer can embed the entire solution into the system by instructing desired joint angle configurations for a robot to follow.
Approaches for more complicated tasks might embed physical modelling into the control system, however this is often brittle because many real-world physics effects are difficult to capture accurately.

In this paper we consider the family of industrial insertion tasks where the robot inserts a grasped part into a tight-fitting socket.
Today, the engineering time for fine-tuning state-of-the-art feedback controllers for such tasks can be similar in cost to the robot hardware itself.
Flexible manufacturing with smaller lot-sizes and faster engineering cycles requires being able to quickly synthesize robust control policies, which can handle variability.
This also broadens the space of manufacturing tasks accessible to robot automation.
Notably, while the family of insertion tasks share significant structure, few existing methods have demonstrated the capability to take advantage of that similarity.
Many of the most effective current methods for compliant robotic insertion \cite{whitney1977force, whitney1982quasi, chhatpar2001insertion, park2013intuitive} require physical models, or else rely on manually-tuned feedback controllers to attain satisfactory performance.

Ideally, the task structure for an insertion task should be automatically inferred from the experience of having solved similar tasks. This insight leads us to meta-reinforcement learning methods, which given experience with a \textit{family} of tasks, adapt to a new task from this family. However, while reinforcement learning (RL) methods can solve a task by learning from data, applying RL in the real world on many tasks is expensive.
To circumvent this problem, we represent a task distribution entirely in simulation. Here, we can control various facets of the environment, samples are cheap, and reward functions are easy to specify. In simulation, we learn the latent structure of the task using probabilistic embeddings for actor-critic RL (PEARL), an off-policy meta-RL algorithm, which embeds each task into a latent space \cite{rakelly2019efficient}. The meta-learning algorithm first learns the task structure in simulation by training on a wide variety of generated insertion tasks. For our family of insertion tasks, the size and placement of the components, parameters of the controller, and magnitude of sensor errors are all randomized, resulting in the policy learning robust and adaptive search strategies based on the latent task code. After training in simulation, the algorithm is then capable of quickly adapting to a real-world task.

In this work, we solve industrial robotic insertion problems by learning the latent structure of the tasks with meta-RL. Concretely, we look at the task of grasping and inserting two parts: a Misumi E-model electrical connector into its socket (one of the most challenging tasks from the IROS 2017 Robotic Grasping and Manipulation Competition \cite{roboticgrasping2017iros}) and a gear on a shaft. We adapt the same policy, which was learned in simulation, to each of the two tasks, despite their distinct physical properties. Moreover, in each task, our method adapts with just 20 trials, significantly fewer than in previous work. We present the robotic system, including a system to account for grasp errors from camera images. Finally, we cover the comprehensive evaluation of our method against both conventional methods and learning-based methods for different degrees of environment variability.

\begin{figure*}[btp]
\centering

    \setlength{\unitlength}{\textwidth}
    \begin{picture}(1, 0.4)(0, 0)
        \put(0,0){\includegraphics[width=0.25\textwidth]{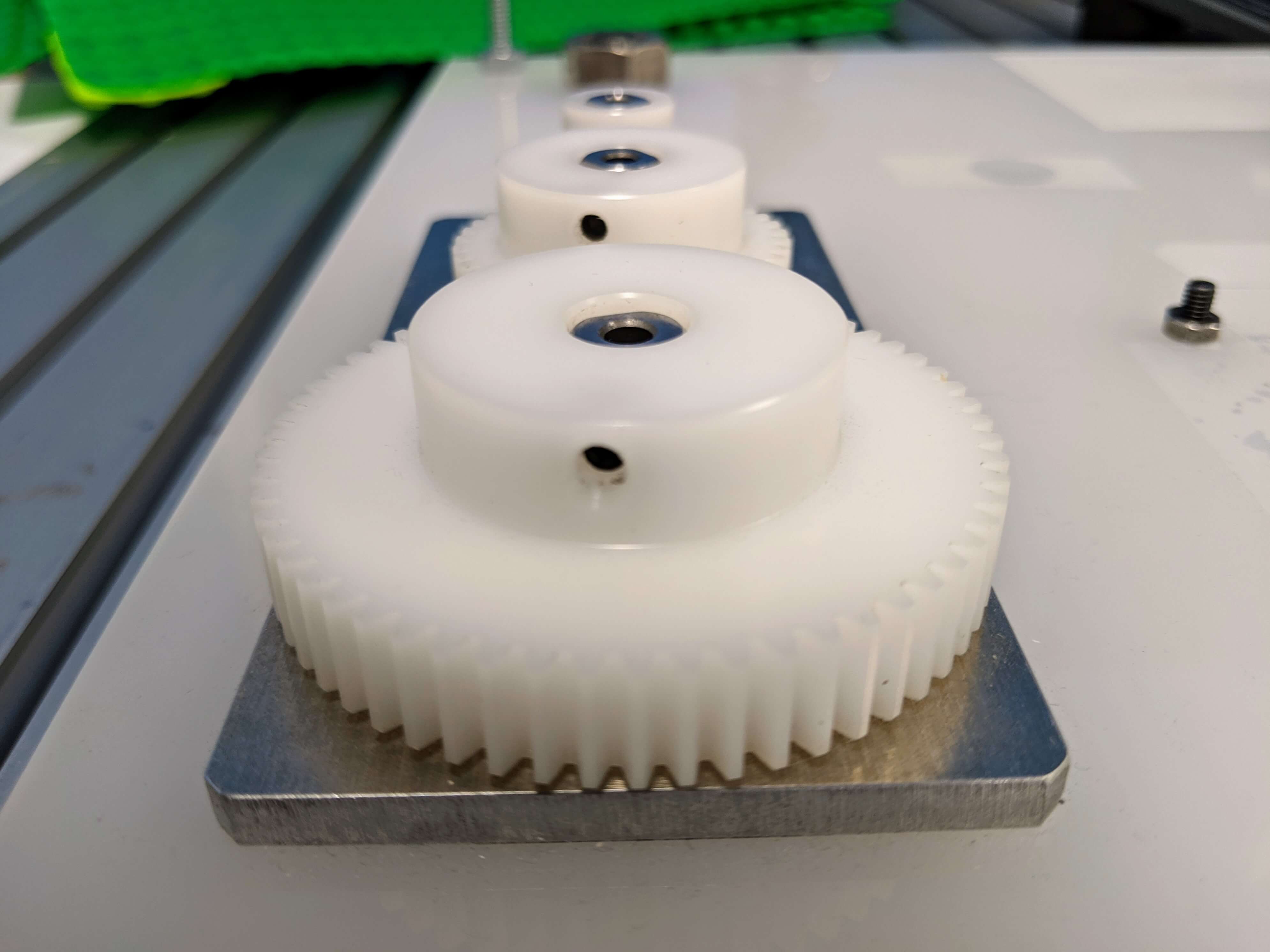}}
        \put(0,0.2){\includegraphics[width=0.25\textwidth]{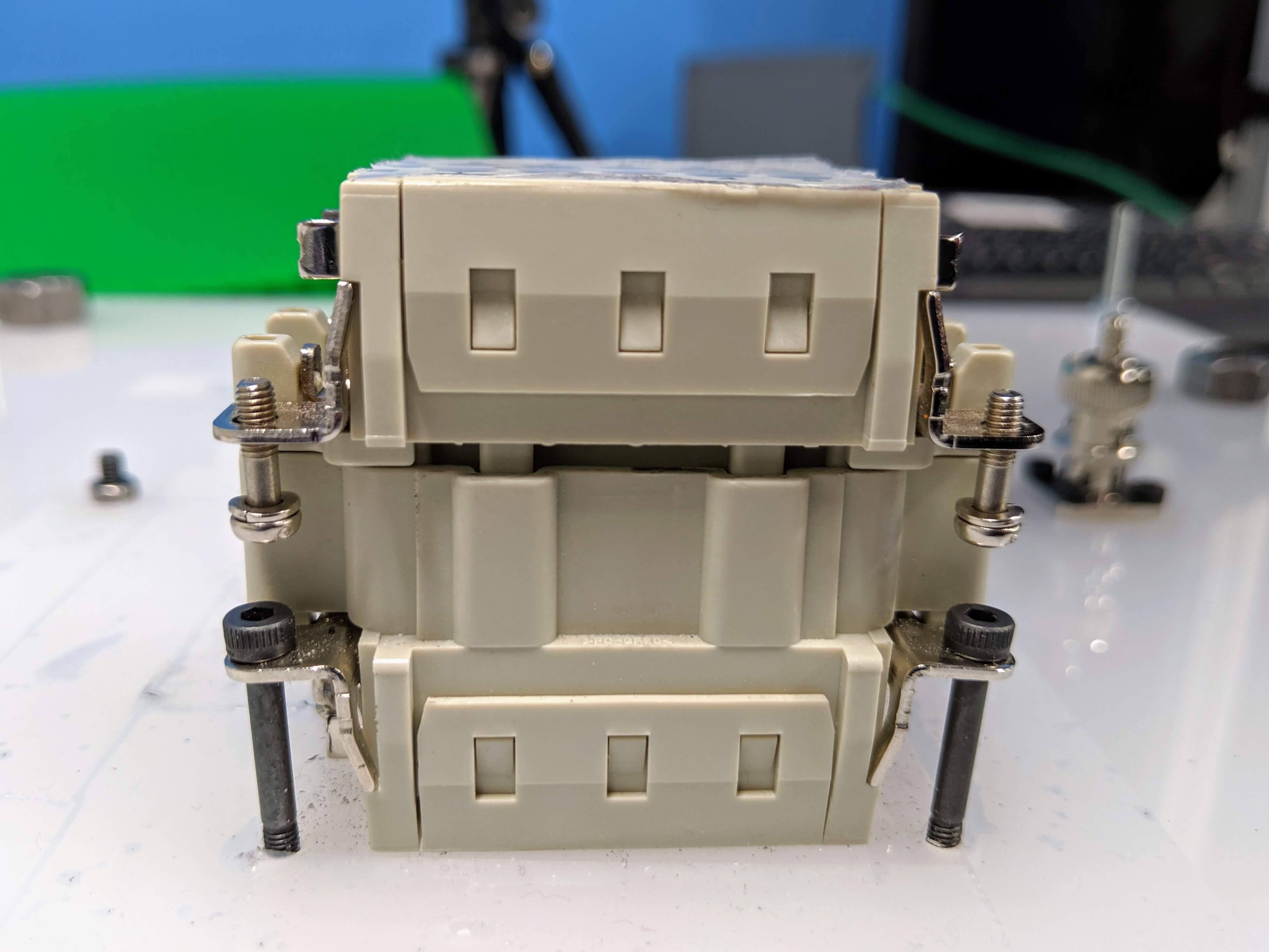}}

        \put(0.26, 0.31){\vector(1, 0){0.15}}

        \put(0.26,0.35){\parbox{1.0in}{1. Design family of simulated tasks \\ $\{\mathcal{T}_1, \mathcal{T}_2, \dots, \mathcal{T}_N \}$ }} 

        \put(0.47, 0.21){\frame{\includegraphics[width=0.2\textwidth]{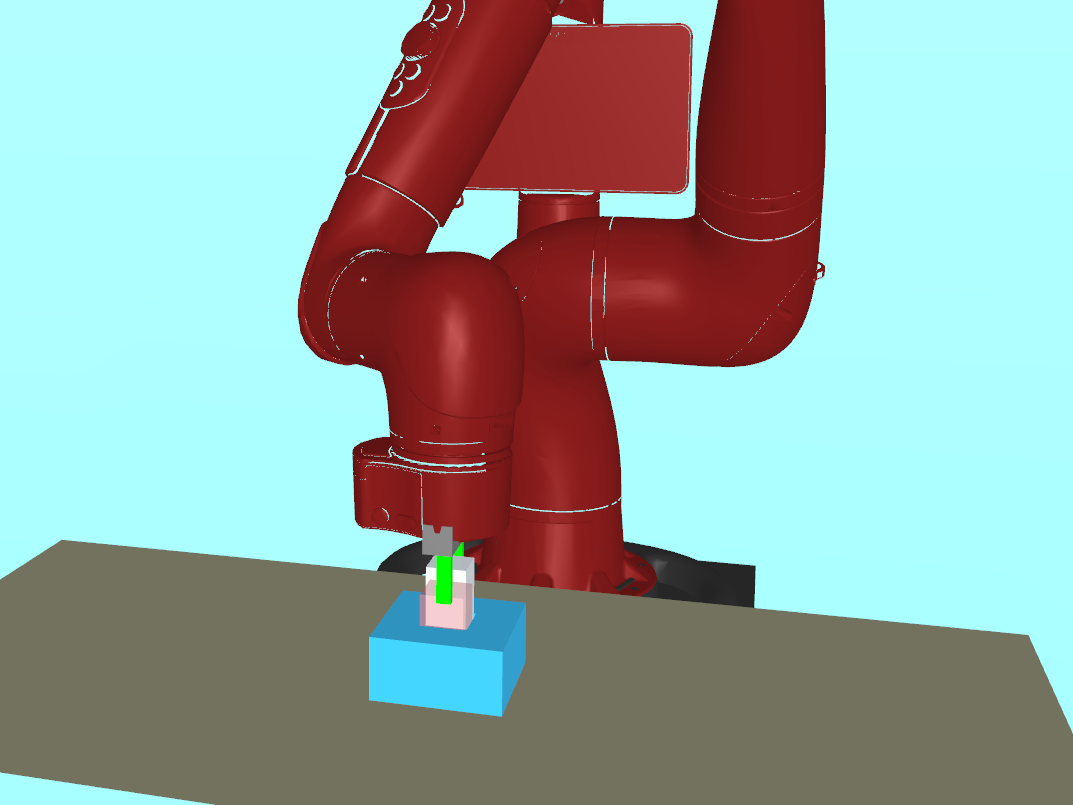}}}
        \put(0.43, 0.17){\frame{\includegraphics[width=0.2\textwidth]{figs/screenshot_sawyer_insertion_simulation_cropped.png}}}
        \put(0.39, 0.13){\frame{\includegraphics[width=0.2\textwidth]{figs/screenshot_sawyer_insertion_simulation_cropped.png}}}
        \put(0.35, 0.09){\frame{\includegraphics[width=0.2\textwidth]{figs/screenshot_sawyer_insertion_simulation_cropped.png}}}
        \put(0.31, 0.05){\frame{\includegraphics[width=0.2\textwidth]{figs/screenshot_sawyer_insertion_simulation_cropped.png}}}
        \put(0.27, 0.01){\frame{\includegraphics[width=0.2\textwidth]{figs/screenshot_sawyer_insertion_simulation_cropped.png}}}

        \put(0.71,0.33){\parbox{0.28\textwidth}{2. Learn latent embedding of task context $q_\phi(z|c)$ and policy $\pi_\theta(a|s, z)$ on family of simulated tasks using PEARL \cite{rakelly2019efficient}}}

        \put(0.53, 0.05){\vector(1, 0){0.15}}
        \put(0.49,0.02){\parbox{0.25\textwidth}{3. Adapt to the real world}}

        \put(0.72, 0.0){\includegraphics[trim=0 0 800 0, clip,  width=0.27\textwidth]{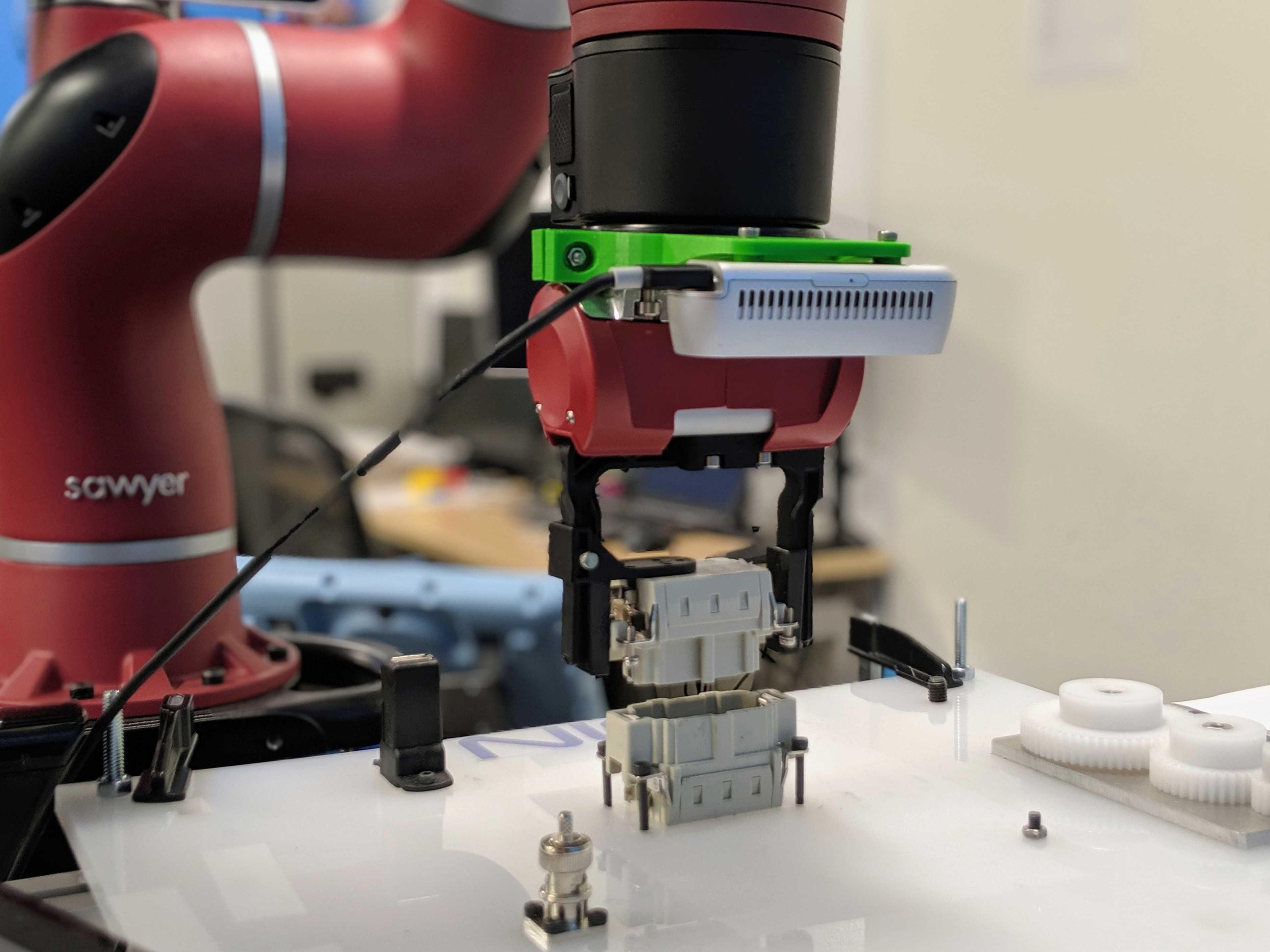}}
    \end{picture}
  \urlstyle{tt}
  \caption{
  We present results on solving two real-world use cases of robotic industrial insertion tasks: plugging in an electrical connector and a gear assembly task, both shown on the left. We model a family of simulated insertion tasks by randomizing simulator parameters. Next, we use meta-reinforcement learning in simulation to learn a latent embedding of tasks and a policy that can rapidly adapt to a new task in that family. Finally, we show that the policy can indeed be quickly adapted to real-world tasks with only 20 trials on the physical robot. Videos and other materials are available at  \url{http://pearl-insertion.github.io}.
  }
\label{fig:title_page_figure}
\end{figure*}

\section{Related Work}\label{sec:relatedwork}

Studies on peg-in-hole insertion have been ubiquitous in industrial automation, as it is representative of many common assembly problems.
The key challenges involved in insertion tasks are modeling of physical contacts, and handling errors in perception and control.
Since physical modeling of contacts and friction is often difficult, deployed controllers for insertions are based on heuristic search patterns that handle the issues implicitly.
These methods include random search, spiral search or raster search~\cite{chhatpar2001insertion}.
The search patterns are combined with compliant control methods, which have been amongst the first model-based strategies for solving insertion tasks \cite{whitney1977force, whitney1982quasi, park2013intuitive}.
The parameters of these controllers are manually tuned in order to overcome perception and control errors for a specific system.
The search patterns are often embedded in control state-machines, which guide the system.
Other approaches focus on high-precision assembly by taking advantage of high-dimensional geometry or contact information \cite{li2014usbgelsight, tamar2017hindsightplan, inoue2017deeprlassembly, luo19variableimpedance}.
These approaches require a significant amount of engineering, modeling and tuning.

Instead of relying on human ingenuity to solve robotic control tasks, the paradigm of RL has promise to autonomously learn the control policy from data. RL has thus far been used in a variety of settings, such as playing ping-pong~\cite{peters2010reps}.
RL with expressive function approximators, or deep RL, further allows tasks to be learned from raw sensor inputs such as images. Deep RL has shown success in games~\cite{mnih2013atari, silver2016alphago}, in learning fine robotic manipulation skills~\cite{levine2016gps} and navigation~\cite{kahn2018navigation}. Specifically, peg insertion tasks have commonly been considered in deep RL settings. Florensa et al. investigate difficult insertion tasks with sparse rewards in simulation using a reverse curriculum \cite{florensa2017resets}. Another approach to solving these tasks is to use prior data such as demonstrations \cite{hester17dqfd, nair2018demonstrations, rajeswaran2018dextrous, schoettler2019insertion}. Vecerik et al. \cite{vecerik17ddpgfd} perform a real-world insertion task utilizing demonstrations. Alternatively, one can learn a residual policy for contacts that is superposed with conventional controls \cite{johannink18residualrl, silver18residualpolicylearning}.

Another line of work considers first learning on simulation models of a task and then transferring the policy to the real world. One approach is domain randomization, which trains on a wide distribution of tasks in simulation assuming that the real world task is captured in that distribution \cite{sadeghi2017simtoreal, tobin2017domainrandomization, peng2018openai, pinto2017asymmetric}. Further work adaptively randomizes the distribution of the tasks \cite{openai2019cube, ramos2019bayessim, mehta2019adr}. One can also actively adapt the simulator by switching between simulation and real-world interaction to guide the simulator \cite{zhou2019epi, chebotar2019simtoreal}. In this work, we take a different approach by using meta-learning to learn a distribution of skills in simulation, followed by adaptation in the real world.

\section{Background}
In this section, we define basic notation and describe reinforcement learning and meta-learning.

\subsection{Reinforcement Learning}
We model our problem as a Markov decision process, where an agent at every discrete timestep $t$ is in state $x_t \in \mathcal{X}$, executes an action $a_t \in \mathcal{A}$, receives a scalar reward $r_t(s_t, a_t)$, and the environment evolves to the next state according to the transition probability $p(x_{t+1}|x_t, a_t)$. The agent attempts to maximize the expected return $R = \sum_{t=0}^H \gamma^t r_t$ where $H$ is the planning horizon and $\gamma \in (0, 1)$ is a discount factor. In reinforcement learning, the agent learns a policy $a_t = \pi_\theta(x_t)$ that is optimized from data.

\subsection{Meta-Learning}
Meta-learning is the problem of rapid adaptation: given experience in some family of tasks, how can use that experience to quickly adapt to a new task at test time? Formally, consider a task $\mathcal{T} = \{r(s_t, a_t), p(x_1), p(x_{t+1}|x_t, a_t)\}$ to be defined by the reward function $r_t(s_t, a_t)$, initial state distribution $p(x_1)$, and transition distribution $p(x_{t+1}|x_t, a_t)$.
We consider some distribution over tasks $p(\mathcal{T})$, which we want to perform well on at test time by collecting limited experience during training time.

Several methods have explored this setting. One class of methods separates the training time into meta-training and meta-testing, and attempts to learn a model (a policy,
forward model, or loss function) during meta-training that improves meta-test performance \cite{finn2017maml, Grant2018, Srinivas2018, bharadhwaj2019simtoreal, wortsman2019learntolearn}. In the meta-RL setting, these methods effectively take advantage of back-propagation through on-policy gradient updates, which limits their sample efficiency.

The other class of methods effectively learn a latent representation of the task \cite{duan2016rl2, andrychowicz2016metalearning, rakelly2019efficient}. The last of these can take advantage of off-policy data, allowing sample-efficient learning on real robots, and we describe the algorithm further in the following section.

\subsection{Probabilistic Embeddings for Actor-Critic RL}

Probabilistic embeddings for actor-critic RL (PEARL) is a meta-learning algorithm that enables sample efficient meta-learning by reusing past data with off-policy RL algorithms \cite{rakelly2019efficient}. The key idea is to condition the policy on the past transitions of the current task, which is termed the context $c^\tau_{1:n}$. The context is encoded into a latent variable $Z$, and we train the policy $\pi_\theta(a|s,z)$. During meta-training PEARL learns the policy parameters and the inference network $q_\phi(z|c)$ which is factorized as $q_\phi(z|c_{1:N})= \prod_{n=1}^N \Psi(z|c_n)$ and $\Psi$ are Gaussian factors, resulting in a Gaussian posterior. The parameters $\theta$ and $\phi$ are learned with an off-policy algorithm that additionally learns a critic. At meta-test time, $z \sim q(z|c)$ is sampled before every rollout, and the new data is used to update the posterior.

\section{Industrial Insertion Tasks}\label{sec:tasks}

We apply meta-learning to two real-world industrial insertion tasks, a waterproof electrical connector plug and a 3D-printed gear, pictured in Fig.~\ref{fig:title_page_figure}.
For our experiments, we use the Rethink Robotics Sawyer robot running a Cartesian-space end effector position controller, further detailed in \ref{sec:impedance_controller}. Thus the action-space is 3-dimensional.
As observations, the current end effector positions relative to the the assumed goal location are used, resulting in a 3-dimensional observation space. Each real-world trial consists of 50 steps with a maximum step size of $2\,\mathrm{mm}$. The duration $\Delta T$ of each step is calculated by multiplying the length of the step with a desired average velocity of $0.01\,\mathrm{m}/\mathrm{s}$. After each trial a reset is performed, the reset position is located $5\,\mathrm{mm}$ above the insertion socket. The workspace of the robot is defined as a cylinder with a radius of $3\,\mathrm{cm}$ and a height of $4\,\mathrm{cm}$, centered at the goal location. If it happens that the robot leaves the workspace, it gets pushed back inside, perpendicular to the nearest surface of the cylinder.
If an insertion is completed before the end of a trial, the end effector is kept still but rewards are collected until the end of the trial.
We use the following sparse reward function during the real-world adaptation phase:
\begin{equation}
 r_t =
  \begin{cases}
   1        & \text{if } \mathrm{height} \leq \mathrm{threshold} \\
   0        & \text{otherwise,}
  \end{cases}
\label{eq:sparse_reward_function}
\end{equation}
where the threshold to detect a successful insertion with a height measurement is $5\,\mathrm{mm}$ below the tip of the insertion.

\section{Method}\label{sec:method}

The key insight of this work is that industrial insertion tasks have shared structure that can be exploited by learning from data on a family of tasks.
Thus, in order to obtain a general meta-RL policy for the real-world, we first design a family of tasks in simulation to reflect the real world tasks.
Then, we use meta-learning in simulation to learn a policy and task embedding that allows fast adaptation to new tasks in that family.
Finally, we apply the learned policy in the real world, where the complete task is to first grasp a part and then insert it. Below, we describe each of these steps in detail.

\subsection{Simulated Environment Design}\label{sec:experiment_setup_sim}

\begin{figure}[hbtp]
      \centering
\begin{subfigure}{0.5\linewidth}
    \centering
    \includegraphics[width=1.0\linewidth]{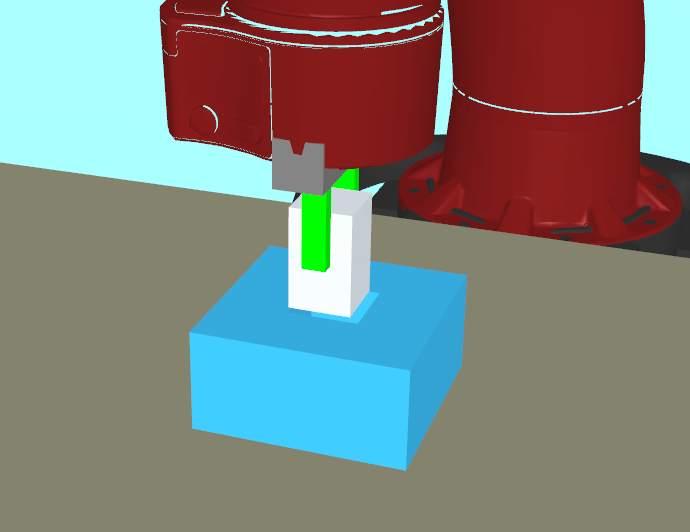}
\end{subfigure}
  \caption{In the simulated environment, we model the connector insertion with a square block that is inserted into a slightly larger square hole. The parameters of the simulator are randomized to generate a family of insertion tasks of different difficulties.}
      \label{fig:simulation_environment}
\end{figure}

To simulate the family of industrial insertion tasks, we use the physics engine MuJoCo \cite{todorov12mujoco}.
The simulated environment, shown in Fig.~\ref{fig:simulation_environment}, contains the Sawyer robot, a table, blocks on the table that form a hole, and a block that fits into this hole located in the robot's parallel gripper.
The blocks on the table are fixed and can not move, and the block inside the robot's gripper is welded to the end effector.
Like the real world, Cartesian-space end effector position control is used, with the maximum step size in each of the 3 directions set to $2\,\mathrm{mm}$.
The family of tasks is generated by randomizing simulation parameters. The following parameters are randomized:
\begin{itemize}
    \item $O$, the horizontal offset of the goal, within $\pm5\,\mathrm{mm}$.
    \item $C$, the clearance of the insertion task, modified by changing the size of the block between $13\,\mathrm{mm}$ and $14\,\mathrm{mm}$, while the size of the square hole remains fixed at $15\,\mathrm{mm}$.
    \item $S$, the scaling of the position controller's step size, in the range $\pm10\%$.
\end{itemize}
Additionally, the reset position of the end-effector is uniformly sampled inside a cube with side length $5\,\mathrm{mm}$, located $5\,\mathrm{mm}$ above the ideal goal, before each reset.

The observation space of the simulated environment consists of the 3-dimensional end effector location, measured relative to the ideal goal, to which the random perturbations are added.
Centering observations with respect to the calibrated goal location allows the reuse a final policy on different robot setups.

The reward function during the simulated meta-training is the $\ell_2$-distance to a full insertion with the current goal location. During the meta-adaptation phase, the sparse reward function in Eq.~\eqref{eq:sparse_reward_function} is used. This is done because the exact goal location is not known during test-time, but a successful insertion can be indicated via a height measurement. The choice of rewards during training and test time are comparable to prior work \cite{rakelly2019efficient}.

\subsection{Sim-to-Real Transfer via Meta Reinforcement Learning}
\label{sec:sim2real_transfer}
Using the simulator, we train a policy with the meta-RL algorithm on the family of tasks.
Although any meta-RL algorithm could potentially be used, in this work we use PEARL for a number of reasons.
First, due its capability for off-policy training, it is highly sample efficient.
Second, PEARL learns a task embedding, which allows it to explicitly learn a latent structure over the family of tasks.
This property of the algorithm also allows for very fast adaptation, which is vital in the real-world as collecting real-world samples can be expensive.
The training of PEARL is outlined in Alg.~\ref{alg:sim_to_real}.
The meta-RL policy trained in simulation is then able to adapt to tasks sampled from the training distribution within a small number of trials.

We then perform policy adaptation on the real system until consistent performance is reached, as detailed in Alg.~\ref{alg:sim_to_real2}.
From the perspective of the algorithm, the real system is just another task to be adapted to.
This simple procedure is surprisingly effective at learning robust, adaptive controllers.

\begin{algorithm}[tbp]
   	\caption{PEARL Training in Simulation}
   	\label{alg:sim_to_real}
   	\begin{algorithmic}[1]
       	\REQUIRE Batch of simulated training tasks $\{\mathcal{T}_i\}_{i=1\dots T}$ from $p(\mathcal{T})$, learning rates $\alpha_1, \alpha_2, \alpha_3$
       	\STATE Initialize replay buffers $\mathcal{B}^i$ for each training task
       	\WHILE{not done}
       	    \FOR{each $\mathcal{T}_i $}
                \STATE Initialize context $\bm{\mathrm{c}}^i = \{ \}$
                \FOR{$k = 1, \dots, K$}
                \STATE Sample $\bm{\mathrm{z}}\sim q_\phi(\bm{\mathrm{z}} \vert \bm{\mathrm{c}}^i)$
                \STATE Gather data from $\pi_\theta(\bm{\mathrm{a}}\vert \bm{\mathrm{s}}, \bm{\mathrm{z}})$ and add to $\mathcal{B}^i$
                \STATE Update $\bm{\mathrm{c}}^i = \{(\bm{\mathrm{s}}_j, \bm{\mathrm{a}}_j, \bm{\mathrm{s}}'_j, r_j)\}_{j:1\dots N}\sim \mathcal{B}^i$
                \STATE Sample $\bm{\mathrm{z}}\sim q_\phi(\bm{\mathrm{z}}\vert\bm{\mathrm{c}}^i)$
                \ENDFOR
            \ENDFOR
            \FOR{step in training steps}
                \FOR{each $\mathcal{T}_i $}
                    \STATE Sample context $\bm{\mathrm{c}}^i \sim \mathcal{S}_c(\mathcal{B}^i)$ and RL batch  ${b^i\sim\mathcal{B}^i}$
                    \STATE Sample $\bm{\mathrm{z}}\sim q_\phi(\bm{\mathrm{z}} \vert \bm{\mathrm{c}}^i)$
                    \STATE $\mathcal{L}_{actor}^i = \mathcal{L}_{actor}(b^i, \bm{\mathrm{z}})$
                    \STATE $\mathcal{L}_{critic}^i = \mathcal{L}_{critic}(b^i, \bm{\mathrm{z}})$
                    \STATE $\mathcal{L}_{KL}^i=\beta D_{\mathrm{KL}}(q(\bm{\mathrm{z}}\vert\bm{\mathrm{c}}^i)\Vert r(\bm{\mathrm{z}}))$
                \ENDFOR
                \STATE $\phi\leftarrow\phi-\alpha_1\nabla_\phi\sum_i(\mathcal{L}_{critic}^i+\mathcal{L}_{KL}^i)$
                \STATE $\theta_\pi\leftarrow\theta_\pi-\alpha_2\nabla_\theta\sum_i\mathcal{L}_{actor}^i$
                \STATE $\theta_Q\leftarrow\theta_Q-\alpha_3\nabla_\theta\sum_i\mathcal{L}_{critic}^i$
            \ENDFOR
        \ENDWHILE
   	\end{algorithmic}
\end{algorithm}

\begin{algorithm}[tbp]
           	\caption{PEARL Sim-to-Real Adaptation}
           	\label{alg:sim_to_real2}
           	\begin{algorithmic}[1]
           	\REQUIRE Trained Meta-RL poliy $\pi_\theta$, trained context encoder $q_\phi$, real test task $\mathcal{T}$
            \FOR{$k=1,\dots,K$}
                \STATE Sample $z\sim q_\phi(\bm{\mathrm{z}}\vert c^\mathcal{T})$
                \STATE Roll out policy $\pi_\theta(\bm{\mathrm{a}}\vert\bm{\mathrm{s}},\bm{\mathrm{z}})$ to collect data $D_k^\mathcal{T}=\{(\bm{\mathrm{s}}_j, \bm{\mathrm{a}}_j), \bm{\mathrm{s}}'_j, r_j\}_{j:1,\dots,N}$
                \STATE Accumulate context $\bm{\mathrm{c}}^\mathcal{T}=\bm{\mathrm{c}}^\mathcal{T} \cup \bm{\mathrm{D}}^\mathcal{T}_k$
            \ENDFOR
       	\end{algorithmic}
\end{algorithm}

\subsection{Real-World Execution}

While we only train the insertion skill in simulation, in the real world the task is to first grasp the part and then insert it. In this section, we cover the real-world implementation details including a more accurate controller for the Sawyer, and an algorithm for grasp detection and correction.

\vspace{0.3cm}

\subsubsection{Robot Impedance Controller} \label{sec:impedance_controller}
The control scheme we developed for precise end effector position control of the Sawyer robot is presented in Alg.~\ref{alg:robot_control}. With this controller, the robot consistently reaches a target with a precision of $0.1\,\mathrm{mm}$. In addition to the low-level control, we added a non-interfering high-level impedance controller, that does not decrease precision. Using position commands instead of velocity commands resulted in an average position error of $0.4\,\mathrm{mm}$. With the default end effector position provided by the manufacturer, a target was reachable within $1\,\mathrm{mm}$, the provided impedance controller showed an error of $10\,\mathrm{mm}$.

\begin{algorithm}[tbp]
           	\caption{Robot Control Scheme}
           	\label{alg:robot_control}
           	\begin{algorithmic}[1]

           	\REQUIRE desired end effector location, orientation, and duration for action $\Delta T$
           	\STATE Calculate desired joint angles $\gamma_{\mathrm{des}}$ via inverse kinematics
           	\STATE Form smooth spline $\bm{\mathrm{S}}$ between $\gamma_0$ and $\gamma_{\mathrm{des}}$
           	\WHILE{$t < t_0+\Delta T$}
           	    \STATE Evaluate $\nabla\gamma_{t}$ as slope of $\bm{\mathrm{S}}$ at time $t$
           	    \STATE Send $\nabla\gamma_{t}$ as joint velocity command to actuators
   	        \STATE Measure end effector forces $\bm{\mathrm{f}}_t$ at $10\,\mathrm{Hz}$
       	        \IF {any $\bm{\mathrm{f}}_t > \bm{\mathrm{f}}_{\mathrm{max}}$}
       	            \STATE break
           	    \ENDIF

            \ENDWHILE
       	\end{algorithmic}
\end{algorithm}

To safely perform insertion tasks, we developed an impedance controller that operates in end effector position space. After each execution of Alg.~\ref{alg:robot_control}, the vertical force at the end effector is measured. If it exceeds a threshold of $ 6\,\mathrm{N}$, a small upwards move is initiated. If the force still exceeds the threshold, a $0.1\,\mathrm{mm}$ larger upwards move follows. This procedure can also be used to achieve a desired downwards force, which we do in experiments with policies that only control the horizontal movement.
An additional safety feature, shown in Alg.~\ref{alg:robot_control}, is a low frequency measurement of the end effector forces in-between the high frequency commands that are sent to the robot. When a threshold of $\bm{\mathrm{f}}_{\mathrm{max}} = 10\,\mathrm{N}$ is exceeded, the robot stops the current motion and waits for the next commanded action. During upwards movements, this safety feature is disabled to prevent the robot from getting stuck while pressing down.

\vspace{0.3cm}

\subsubsection{Grasp Algorithm}\label{sec:grasp_algo}

A RealSense D435 depth camera is mounted to the robot arm and used to scan the workspace to calculate a grasp based on a depth image. We clean the depth image from artifacts and use a hand-tuned distance threshold to binarize the image. In most cases, this already extracts individual objects sufficiently. We then apply a contour fining algorithm to extract rectangular contours, check if the size of a found contour matches the assumed object size, and use temporal filtering to average the object location over multiple frames. The grasp will be planned along one of the principal axes of the rectangular bounding box. Hand-eye calibration is used to find the corresponding real-world coordinates in the robot frame.
The requirements for this grasp approach are that the graspable object is clearly detectable in the depth image and that the distance threshold and the assumed object size are set appropriately.

\begin{figure}[tbp]
      \centering
\begin{subfigure}{1.0\linewidth}
    \centering
    \includegraphics[width=1.0\linewidth]{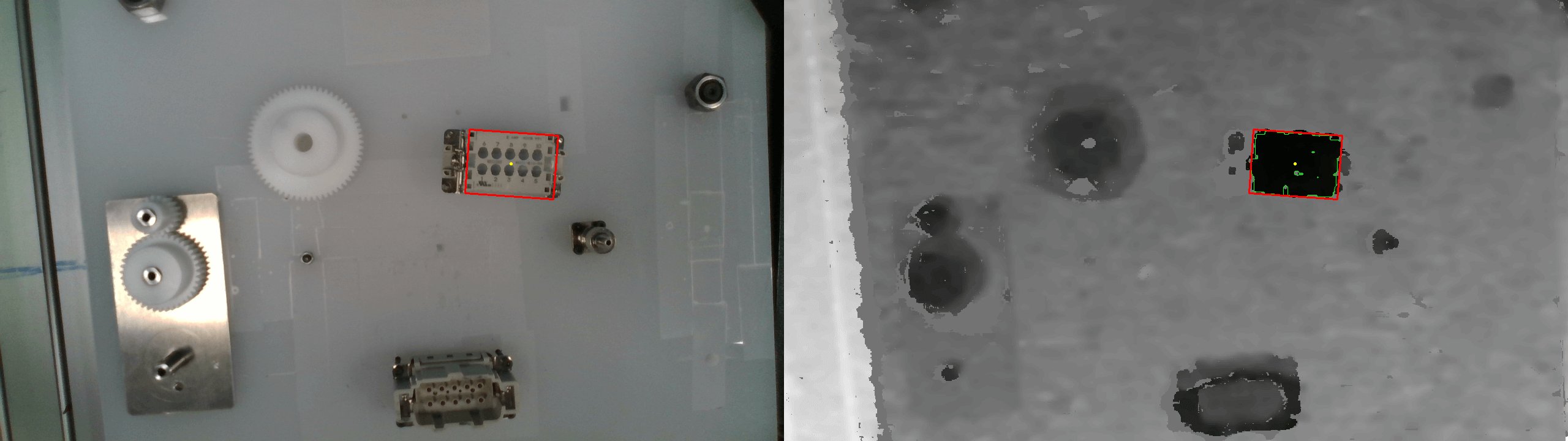}
\end{subfigure}

  \caption{Robot view through a RealSense D435 camera, mounted to the robot arm. The depth image is binarized with a tuned threshold and contours are extracted. A rectangular bounding box (red) is drawn for contours of the expected size and used to calculate an aligned and centered grasp. Averaging the calculated grasps over over multiple frames improved the stability, resulting in 100 consecutive successful grasps when only a single object is in frame.}
      \label{fig:grasp_view}
\end{figure}

\vspace{0.3cm}

\subsubsection{Grasp Error Correction}\label{sec:grasp_correction}

In a real factory setting, each object that is about to be inserted by a robot needs to be grasped first.
This increases the time per insertion attempt and induces unavoidable grasp errors when using a non-self-centering parallel gripper.
In order to resemble the real setting as precisely, as possible, we include the grasping in our experimental setup.  To mitigate grasp errors, we propose a grasp-correction algorithm that only requires a single image taken of the bottom of the grasped object to calculate the object's displacement with respect to a reference grasp.

Our grasp correction algorithm uses an image of the bottom of the grasped object and compares it with a reference image using cross-correlation. From the cross-correlation of the new image with the reference image, the translation with respect to the reference grasp pose can be inferred reliably. The goal location is then adjusted based on the computer grasp error.

Rotational grasp errors are not considered because a the objects were not seen to rotate inside the parallel gripper and the rotation of the fixed goal location was calibrated. In different setups however, rotations may be a major source of error and should be investigated.

\begin{figure}[btp]
      \centering
\begin{subfigure}{1.0\linewidth}
\centering
 \includegraphics[width=0.42\linewidth]{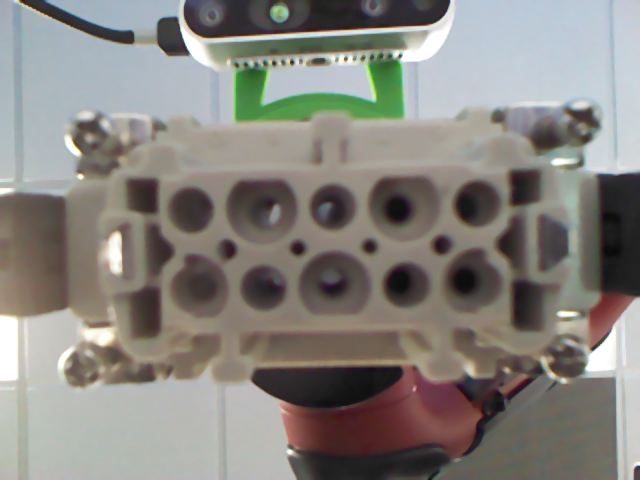}
 \hfil
\includegraphics[width=0.42\linewidth]{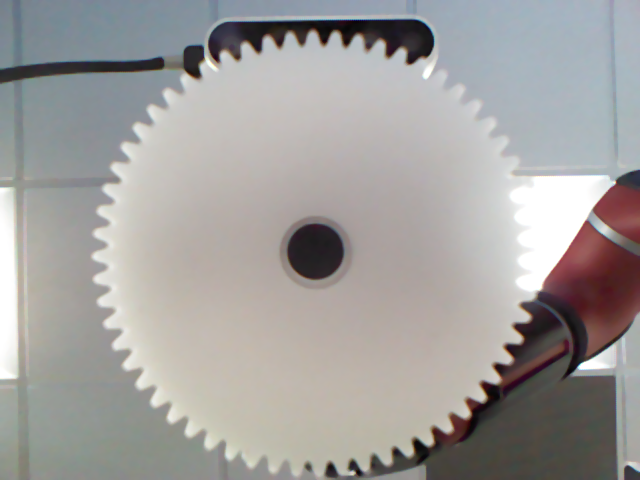}
\end{subfigure}
  \caption{Camera view when scanning the bottom of the part to calculate the pixel offset with respect to a reference grasp. A Kinect~v1 camera is mounted upside down on the table to take an image after each grasp. }
      \label{fig:grasp_photos}
\end{figure}

\section{Experimental Results}\label{sec:experiments}

We conduct a series of experiments to answer the following questions:

\begin{enumerate}[A.]
	\item Can PEARL learn to robustly adapt to novel insertion tasks in simulation?
	\item Is it possible to adapt insertion policies learned using meta-RL in simulation to the real reward?
	\item How does sim-to-real meta-RL compare to existing solutions to robotic insertion problems, in terms of robustness and efficiency?
	\item What patterns and behavior does the algorithm learn in simulation that allow it to transfer to the real world?
\end{enumerate}

We address each of these questions in our experimental evaluation, presented below.

\subsection{Adaptation in Simulation}

\begin{figure}[h]
\centering
    \includegraphics[width=0.90\linewidth]{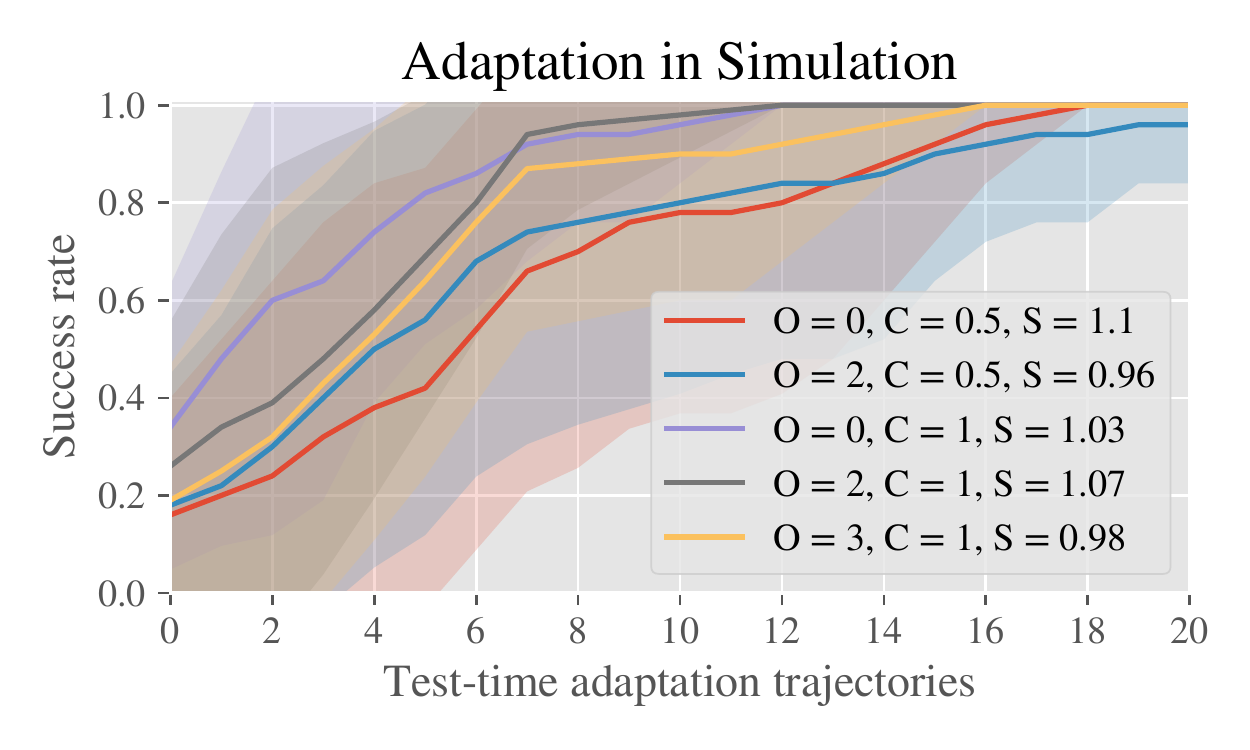}
  \caption{Success rate on test tasks in the adaptation phase in simulation after training. Per experiment, 20 random seeds are evaluated. We see that the PEARL policy successfully learned to adapt to unseen task in simulation. Descriptions of the randomized parameters are given in Section~\ref{sec:experiment_setup_sim}. }
\label{fig:sim_adaptation}
\end{figure}

First, we examine the performance of PEARL on our family of simulated insertion tasks. The adaptation performance on test tasks after training is shown in Fig \ref{fig:sim_adaptation}.

In the results, we see that the zero-shot performance of the trained policies is about $20\%$. But given 20 trials in the new environment, the algorithm can successfully adapt to solve each of the new tasks.

\subsection{Real-World Adaptation}
After training in simulation, we adapt the meta-RL policy to tasks in the real world. As discussed in Sec.~\ref{sec:grasp_correction}, in the real-world tasks the object (either the connector or the gear) is picked up using our grasp system, each grasp is evaluated using a camera image, and grasp errors are compensated according to our grasp correction algorithm.
Since each grasp is slightly different, grasping introduces an additional challenge, requiring our method to compensate for this realistic source of variability.

In addition to the grasping, we consider robustness to poorly calibrated setups by perturbing the goal location. Thus, we evaluate the method on five different tasks between the two use cases: the plug insertion task with no noise, $\pm2$mm noise, and $\pm3$mm noise, and the the gear task with no noise, and $\pm2$mm noise.

\begin{figure}[h]
  \includegraphics[width=0.90\linewidth]{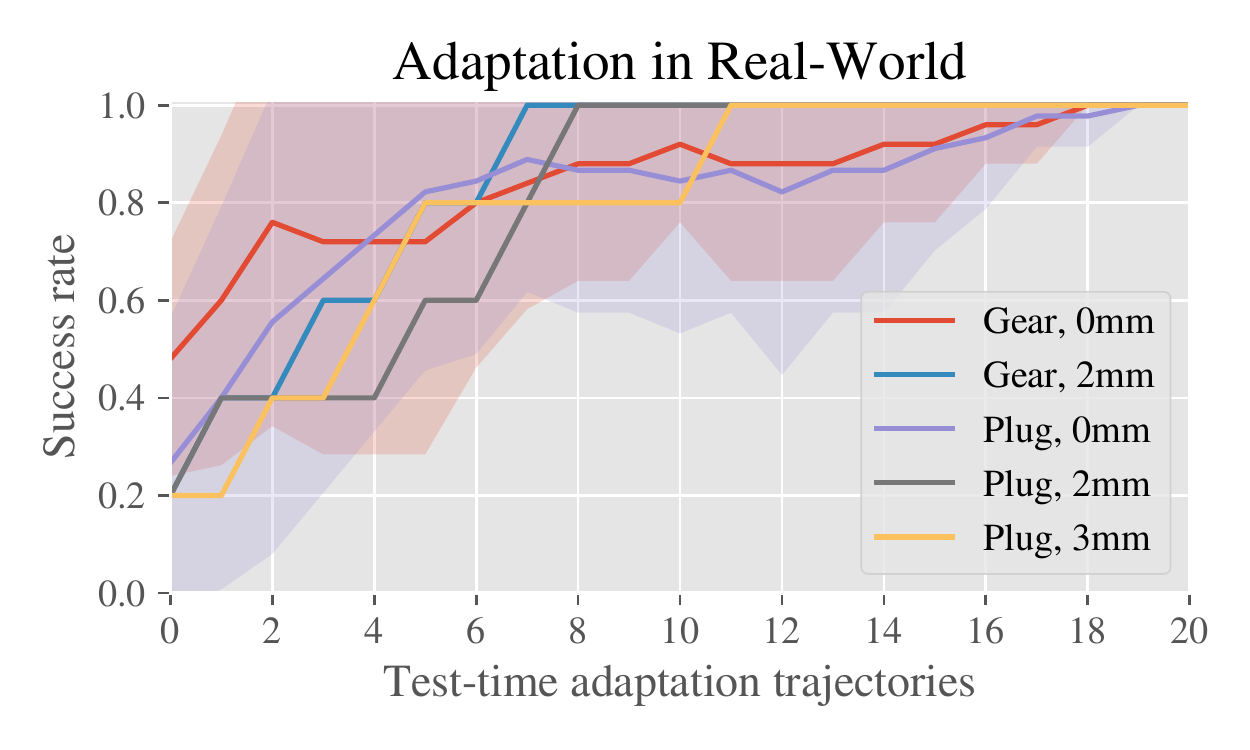}
  \caption{Comparison of the success rates after each trial of the adaptation phase in the real world. Note that in each of the five tasks, our method is able to adapt to the task with less than 20 real-world trials. }
      \label{fig:adaptation_phase_sim_real}
\end{figure}

The real-world adaptation results are presented in Fig.~\ref{fig:adaptation_phase_sim_real}. The results show that in each case we can adapt to all the tasks in less than 20 trials of real-world interaction.

The goal perturbation presents a challenge to all methods. We see that our methods is able to still solve the insertion task. In contrast to the heuristic methods, our method has the capacity to learn very complex search strategies and continuously adapt.
Since the simulated meta-training phase included sufficient randomization of the goal location, the meta-trained policy explores quite broadly when adapting to the real world.

\subsection{Comparison of Robustness and Sample Efficiency}

\begin{table}[tbp]
\renewcommand{\arraystretch}{1.4}
\centering
\caption{Comparison of real-world insertion performance. Moving straight down to the goal location is not a reliable insertion strategy for this task. Due to the high precision requirements, random insertion strategies also fail. Only the policy trained with PEARL can achieve a very high success rate on this task. In the second table, we show average time for insertion in seconds, measurement started at tip of insertion and stopped when fully inserted. }

\vspace{0.2cm}
{ \normalsize
Success Rate
}
\vspace{0.2cm}

\begin{tabular}{l|c|c|c|c|c}
& \multicolumn{3}{c|}{Plug} & \multicolumn{2}{c}{Gear} \\
Policy & \hspace{0.16cm} 0 \hspace{0.16cm} & $\pm2$mm & $\pm3$mm & \hspace{0.16cm} 0 \hspace{0.16cm} & $\pm2$mm  \\
\hline
Straight Down & 0.88 & 0.0 & 0.0 & 0.32 & 0.0\\
\hline
Random Search &  \textbf{1.0} & \textbf{1.0} & \textbf{1.0} & 0.48 & 0.32 \\
\hline
Spiral-Search&  \textbf{1.0} &  \textbf{1.0} & \textbf{1.0} & 0.84 & 0.8\\
\hline
RL from Scratch &   \textbf{1.0} &  0.32 & 0.0 & \textbf{1.0} & 0.92\\
\hline
PEARL Sim2Real & \textbf{1.0} & \textbf{1.0} & \textbf{1.0}& \textbf{1.0} & \textbf{1.0} \\
\end{tabular}

\vspace{0.4cm}
{ \normalsize
Insertion Time
}
\vspace{0.2cm}

\begin{tabular}{l|c|c|c|c|c}
& \multicolumn{3}{c|}{Plug} & \multicolumn{2}{c}{Gear} \\
Policy &  0 & $\pm2$mm & $\pm3$mm & 0 & $\pm2$mm  \\ \hline
\multirow{2}{*}{Straight Down} & $\textbf{3.3}$ & $-$ & $-$ & $\textbf{5.3}$ & $-$ \\
& $\pm0.6$ & & & $\pm1.4$ & \\
\hline
\multirow{2}{*}{Random Search} & 5.6 & $\textbf{7.0}$ & $\textbf{9.7}$ & 20.3 & $23.3$ \\
& $\pm2.0$ & $\pm2.3$ & $\pm4.6$ & $\pm7.3$ & $\pm9.3$ \\
\hline
\multirow{2}{*}{Spiral-Search} &  $6.0$ &  $13.6$ & $26.6$ & $8.0$ & $17.3$ \\
& $\pm4.7$ & $\pm6.6$ & $\pm4.8$ & $\pm2.6$ & $\pm5.2$ \\
\hline
\multirow{2}{*}{RL from Scratch} &  $11.7$ & $-$ & $-$ & $\textbf{5.8}$ & $-$\\
& $\pm 4.2$ & & & $\pm 0.7$ & \\
\hline
\multirow{2}{*}{PEARL Sim2Real} & $5.3$ & $\textbf{6.8}$ & $\textbf{8.2}$& $\textbf{5.7}$ & $\textbf{8.0}$ \\
& $\pm2.7$ & $\pm4.7$ & $\pm4.5$ & $\pm2.6$ & $\pm5.5$ \\
\end{tabular}
\label{tab:success_rates}
\end{table}

In these experiments, we evaluate whether meta-RL is a viable solution for industrial insertion tasks, comparing the method to existing solutions that are used today. We compare to four strong baselines, which are either covered in past research or used widely in industry. In total, we compare the following methods:
\begin{enumerate}
    \item \textbf{Straight downwards.} Move straight downwards from the reset position.
    \item \textbf{Random search.} In this stochastic search policy, described in \cite{marvel2018insertionsearch}, the robot moves horizontally between search points that are sampled uniformly inside of a square shaped search space with side length $6\,\mathrm{mm}$, centered above the assumed goal location. The robot moves downwards at the first sampled point until a vertical contact force of $3\,\mathrm{N}$ is sensed at the end effector. If no successful insertion is detected, the end effector moves back upwards until the measured force decreases below $3\,\mathrm{N}$ and then moves horizontally to the next sampled point where it attempts the another insertion in the same way. At most 50 random insertion attempts are executed in each trial.

    \item \textbf{Spiral search.}
    The robot generates a spiral above the assumed goal location and iteratively attempts to insert downwards at points in the spiral \cite{marvel2018insertionsearch}. During the downwards movement, a force threshold of $3\,\mathrm{N}$ is used to indicate contact and signals the robot to move to the next point in the same way as described above in random search. In our implementation, the distance to the center increases by $0.5\,\mathrm{mm}$ each rotation and insertions are attempted at points  $45\deg$ apart along the spiral. At most, the robot moves through 50 points.

    \item \textbf{RL from scratch.}
    We train SAC \cite{haarnoja2018sac} in the real world from scratch, using the same action space, state space and sparse reward function as in the real-world adaptation phase with PEARL, described in Sec.~\ref{sec:tasks}. The training with SAC requires substantially more environment steps than adapting PEARL, which is why we choose to rigidly mount the adapter to the robot's gripper and leave out the grasping during the training. At test-time, the grasping is performed. The SAC policies were trained for 2 hours and 20 minutes; repeating success was already visible after 1 hour and 20 minutes of training.
    \item \textbf{PEARL Sim2Real.}
    Our method using meta RL, as described in Section \ref{sec:method}.

\end{enumerate}

We evaluate these methods along two dimensions. Most importantly, we measure the success rate of the method on a task. We also measure the time needed for each insertion, to compare the efficiency of the different methods - the moment of successful insertion is detected via a height threshold. The measurement of efficiency is important for practical applications, since throughput is a major consideration in industrial settings. The results of the insertion time were averaged over 10 successful insertions per task and policy.

Results of the experiments performed on all five tasks are reported in Table ~\ref{tab:success_rates}. We immediately see that our method is the only one that consistently solves every task, and is almost always the fastest, except when moving straight down works. The gear use-case is visibly more difficult and not solvable with na\"{i}ve downwards movement. The two heuristic search methods: random search and spiral search, are not always able to succeed at the more difficult settings in the given 50 steps. Meta-RL sim-to-real transfer shows the best performance among the most difficult tasks. Videos of our results can be viewed at \url{http://pearl-insertion.github.io}.

\subsection{What behavior does the policy transfer from simulation?}

\begin{figure}[btp]
\centering
\begin{subfigure}{1.0\linewidth}
    \centering
    \includegraphics[width=0.8\linewidth]{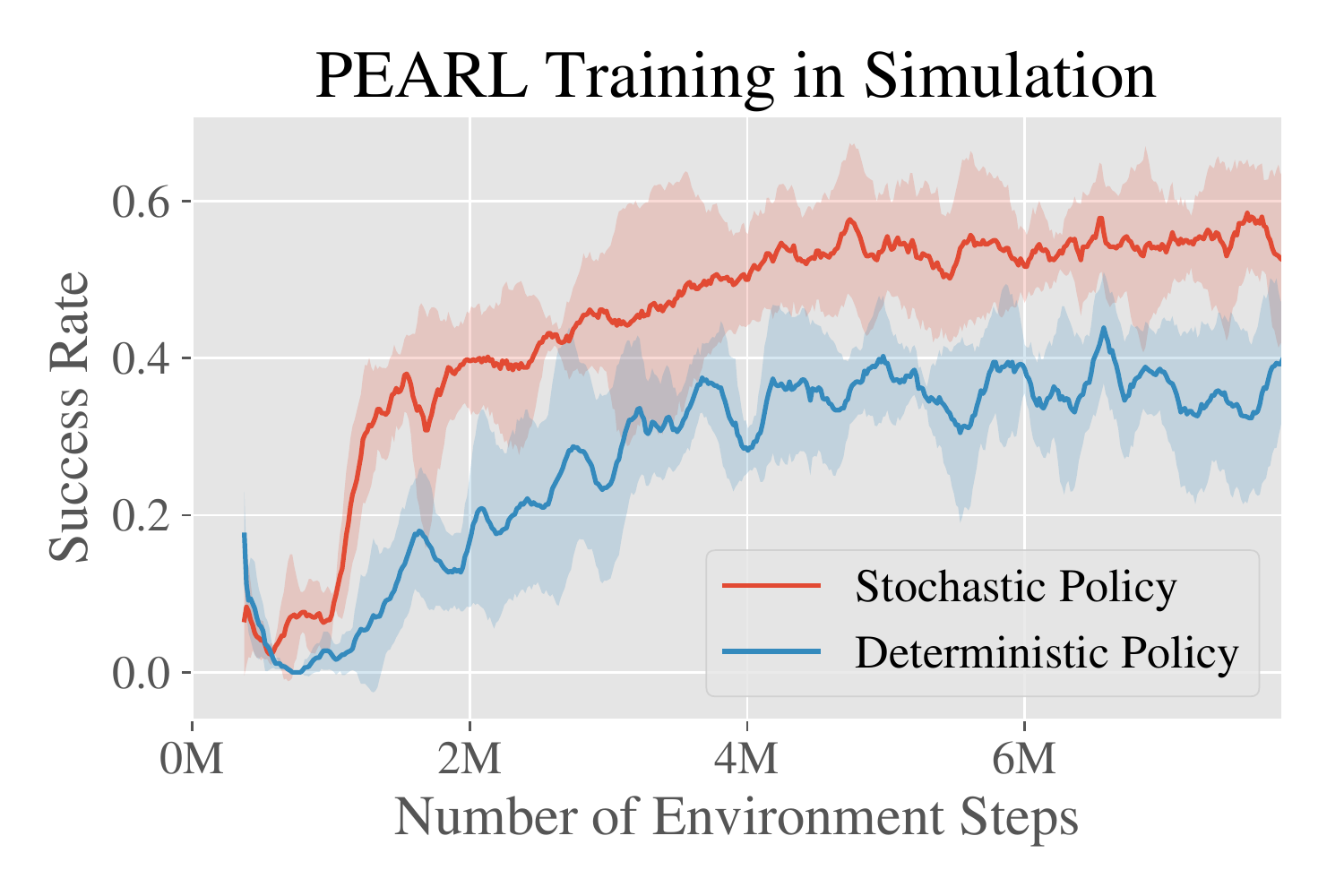}
\end{subfigure}
\hfil
  \caption{Comparison of the PEARL training in simulation with either a deterministic or stochastic policy. Although the deterministic policy is formed by computing the maximum likely action of the stochastic policy, our model of the insertion task benefits from a stochastic policy evaluation. We think this is due to the more extensive exploration during the adaptation step of PEARL.
  }
      \label{fig:det_sto_comparison_plot}
\end{figure}

We believe the main knowledge transferred from sim-to-real is \textit{structured exploration noise}. We investigate by comparing the learned stochastic policy in PEARL to the deterministic evaluation of this policy done by always choosing the most likely action, which is the mean of the output with a Gaussian distribution. Prior work has consistently found that, although stochasticity helps at training time, the deterministic policy gives better final returns \cite{haarnoja2018sac}. In Fig.~\ref{fig:det_sto_comparison_plot} and \ref{tab:det_vs_sto_evaluation}, we compare the stochastic and deterministic policy when learning in simulation and performing sim-to-real transfer with PEARL.

\begin{table}[btp]
\renewcommand{\arraystretch}{1.4}
\centering
\caption{Comparison of deterministic and stochastic evaluation. It was observed that a deterministic policy is equally able to adapt to the real-world setting, but shows slightly less consistency at test-time. When getting stuck at the tip of the insertion, the deterministic policy predominantly failed to recover, whereas the stochastic policy managed to still solve the task in most cases.}
\begin{tabular}{|l|c|c|}
\hline
PEARL Success Rate & Deterministic & Stochastic  \\ \hline
Connector Plug +3mm &  0.44 & 1.0  \\ \hline
Gear +2mm &  0.84 & 1.0 \\ \hline
\end{tabular}
\label{tab:det_vs_sto_evaluation}
\end{table}

\begin{figure}[btp]
\centering
\begin{subfigure}{0.45\linewidth}
    \centering
    \includegraphics[width=1.0\linewidth]{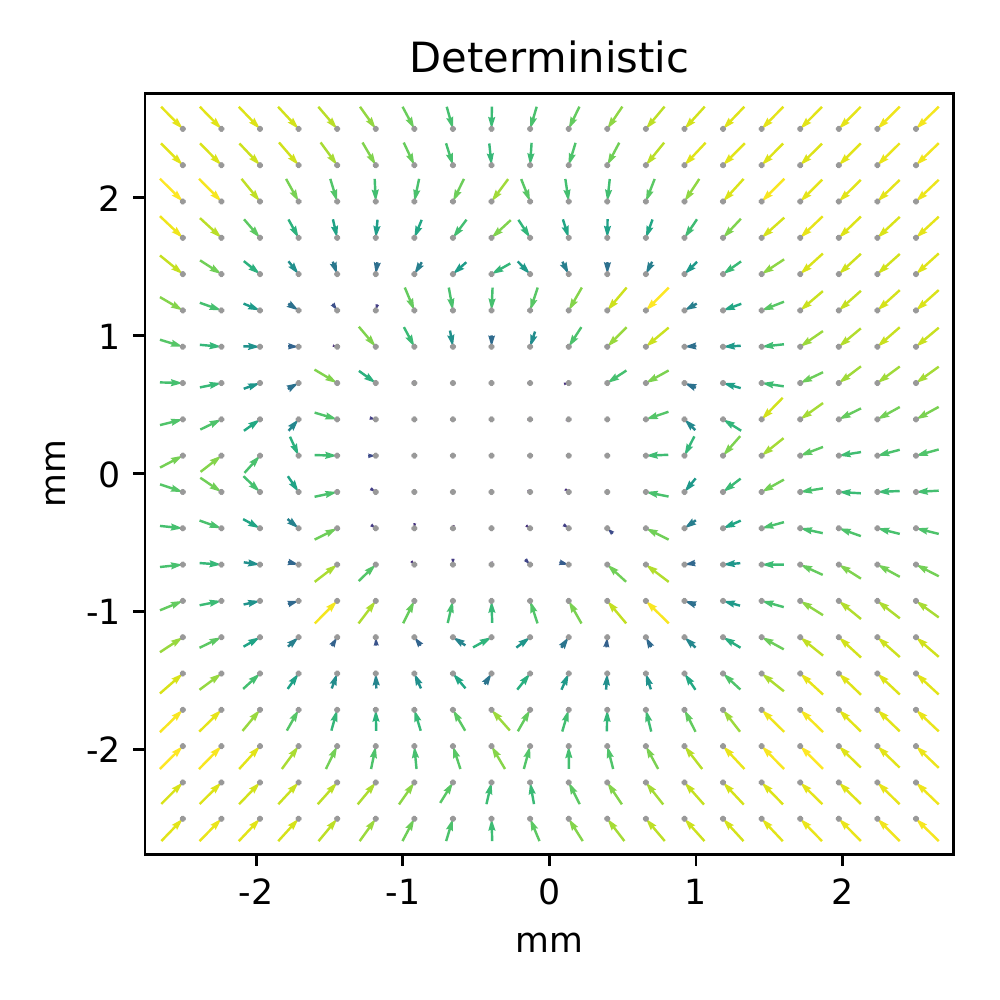}
\end{subfigure}
\hfil
\begin{subfigure}{0.45\linewidth}
    \centering
    \includegraphics[width=1.0\linewidth]{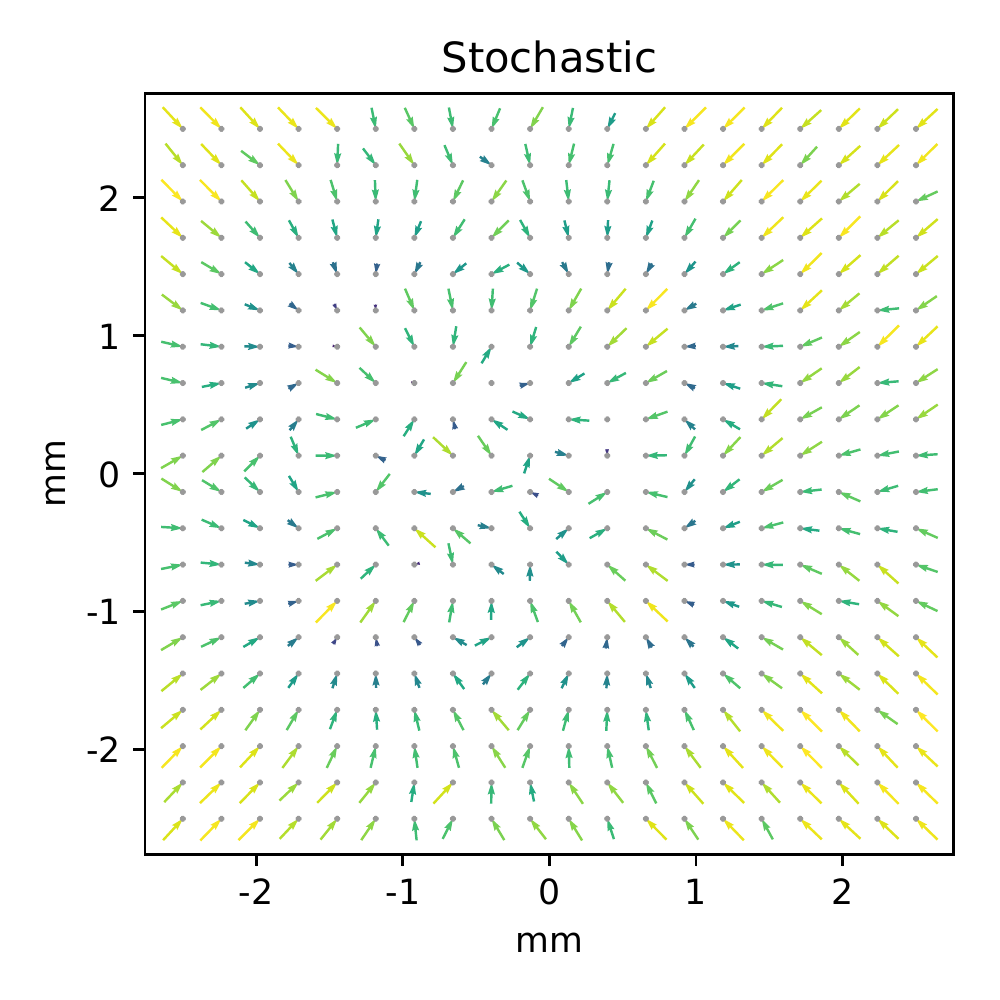}
\end{subfigure}
  \caption{Visualization of the policy outputs on a grid around a goal that is slightly shifted to the left. Here, a policy was trained on a two-dimensional version of our simulated environment, where movement in z-direction was disabled.
  In contrast to a deterministic policy evaluation, a stochastic evaluation also shows movement inside of the goal region, which we observed as beneficial when executing such policy on a real robot.}
  \label{fig:det_sto_adaptation_grid}
\end{figure}

As shown in Fig.~\ref{fig:det_sto_comparison_plot}, the stochastic policy consistently achieves a higher success rate. During the real-world adaptation, we observed better exploration with the stochastic policy, as well as a slightly better final performance, reported in Tab.~\ref{tab:det_vs_sto_evaluation}. The failed insertion attempts of the deterministic policy happened because the gear became stuck at the first stage of the insertion. This physical phenomena was not modeled in the simulation. However, the stochastic policy was still able to recover in all cases because it produced oscillating movements around the contact point of the insertion. In Fig.~\ref{fig:det_sto_adaptation_grid} we visualize the computed actions of a PEARL policy that was trained on a 2D sparse point robot environment with uniformly distributed goals around the origin and adapted in the real world on the electrical connector plug task. It is visible that the deterministic policy does not perform any movement inside of the goal region, whereas the stochastic policy learned to fully explore the goal region. We observed this movement inside of the goal region to be beneficial when performing insertions in the real world, as a slight misjudgement of the shape, size and location of the goal region can be compensated with these stochastic actions.

Finally, we can infer what behavior is learned by analyzing the situations in which the sim-to-real transfer with PEARL did not work well. For instance, the real-world adaptation failed when the randomization of the reset position was left out during the training in simulation. The trained meta-policy did not learn a stable behavior outside of the direct paths to the training goals. In the real world adaptation phase, inaccuracies of the real robot's movement caused the end effector to enter unstable regions, in which a continuous movement in a direction away from the origin occurred. The real-world adaptation also failed when the randomization amount was too high, as sometimes none of the insertion attempts during the real-world adaptation phase succeeded. Due to the use of sparse rewards, PEARL does not obtain any explicit information about the goal location in this case. When we observed this failure case, we reduced the amount of randomization in simulation.

\section{Conclusion}\label{sec:discussion}
In this paper, we studied meta-reinforcement learning for industrial insertion tasks.
Our method first performs meta-training in a low-fidelity simulation, and then actively adapts to a variety of real-world insertion and assembly tasks.
This approach can solve complex real-world tasks in under 20 trials, performing connector assembly and a 3D-printed gear insertion task.
We also demonstrated the feasibility of our method under challenging conditions, such as noisy goal specification and complex connector geometries.

Our method shifts the burden of engineering robotics solutions from designing accurate analytic physical models to designing a family of representative simulated tasks.
Furthermore, as our method requires experience in the real world only for the final adaptation step, the work of designing the simulation may be amortized across many tasks. Thus, we believe that our work illustrates the potential of meta-RL to provide a scalable and general method for rapid adaptation in manufacturing and industrial robotics.

\section{Acknowledgements}
This work was supported by the Siemens Corporation,
the Office of Naval Research under a Young Investigator
Program Award, and Berkeley DeepDrive.

\addtolength{\textheight}{-12cm}


{ \small
\bibliographystyle{IEEEtran}
\bibliography{IEEEfull}
}

\end{document}